%% file: main.tex
\documentclass[runningheads]{llncs}

\usepackage{graphicx}
\usepackage[utf8]{inputenc}
\usepackage{amsmath}
\usepackage{soul}
\usepackage{multirow}
\usepackage{enumitem}
\usepackage{color}
\usepackage{tabu}
\usepackage{xspace}
\usepackage[hidelinks]{hyperref}
\usepackage{balance}
\usepackage[T1]{fontenc}
\usepackage{booktabs}
\usepackage{textcomp}
\usepackage{dblfloatfix} 
\usepackage{microtype,algpseudocode, algorithm, algorithmicx}
\usepackage{ccicons}
\usepackage{svg}
\usepackage{colortbl}
\usepackage{todonotes}
\usepackage{subfig}
\graphicspath{ {./figures/} }

\begin{document}
\newcommand{\SystemName}{FDFtNet}

\makeatletter
\makeatother

\title{\SystemName: Facing Off Fake Images using Fake Detection Fine-tuning Network}

\author{Hyeonseong Jeon \and
Youngoh Bang \and
Simon S. Woo}

\institute{Department of Artificial Intelligence \\ Sungkyunkwan University, Suwon, S. Korea \\
\email{\{cutz, byo7000, swoo\}@g.skku.edu}
}

\maketitle
\input{abstract.tex}
\input{intro.tex}
\input{related.tex}
\input{design.tex}
\input{result.tex}
\input{ablation.tex}
\input{conc.tex}
\input{ack.tex}

\bibliography{references}
\bibliographystyle{splncs04}

\end{document}

%% file: abstract.tex
\begin{abstract}
Creating fake images and videos such as ``Deepfake'' has become much easier these days due to the advancement in Generative Adversarial Networks (GANs). Moreover, recent research such as the few-shot learning can create highly realistic personalized fake images with only a few images. Therefore, the threat of Deepfake to be used for a variety of malicious intents such as propagating fake images and videos becomes prevalent. And detecting these machine-generated fake images has been more challenging than ever.

In this work, we propose a light-weight robust fine-tuning neural network-based classifier architecture called~\textit{Fake Detection Fine-tuning Network} (\textit{\SystemName}), which is capable of detecting many of the new fake face image generation models, and can be easily combined with existing image classification networks and fine-tuned on a few datasets. In contrast to many existing methods, our approach aims to reuse popular pre-trained models with only a few images for fine-tuning to effectively detect fake images. The core of our approach is to introduce an image-based self-attention module called \textit{Fine-Tune Transformer} that uses only the attention module and the down-sampling layer. This module is added to the pre-trained model and fine-tuned on a few data to search for new sets of feature space to detect fake images. We experiment with our \SystemName~on the GANs-based dataset (\textit{Progressive Growing GAN}) and Deepfake-based dataset (\textit{Deepfake} and \textit{Face2Face}) with a small input image resolution of 64$\times$64 that complicates detection. Our \SystemName~achieves an overall accuracy of 90.29\% in detecting fake images generated from the GANs-based dataset, outperforming the state-of-the-art.
\keywords{Fake Image Detection \and Neural Networks \and Fine-tuning}
\end{abstract}

%% file: intro.tex
\section{Introduction}
The emergence of Generative Adversarial Networks (GANs) \cite{goodfellow2014generative}, which produces high-quality images through a generator and a discriminator that is trained adversely and competitively, enables the generated outputs to be highly realistic and sophisticated~\cite{karras2017progressive,zakharov2019few,karras2019style,wu2019sliced}.
However, such high-quality images and videos generated by machines have been abused and harmed the general public (e.g., DeepFake~\cite{deepfake}). Furthermore, a recent study using the few-shot learning technique~\cite{sun2019meta} in GAN allows Deep Learning models to produce high-quality outputs with only a small amount of training data. Zakharov et al.~\cite{zakharov2019few} demonstrated that models capable of generating highly realistic personalized talking head faces could be constructed using few-shot learning techniques, where the training inputs provide attention to the generator as a compressed form of feature landmarks, extracted through embedding layers. Leveraging this method, DeepFake can easily be generated even with only a small amount of training data. Recently reported incidents~\cite{yin_2019} related to DeepFake~\cite{deepfake} and DeepNude show that these technologies are an imminent threat to the public.

\begin{figure}[t]
    \centering
     \includegraphics[width=0.7\linewidth]{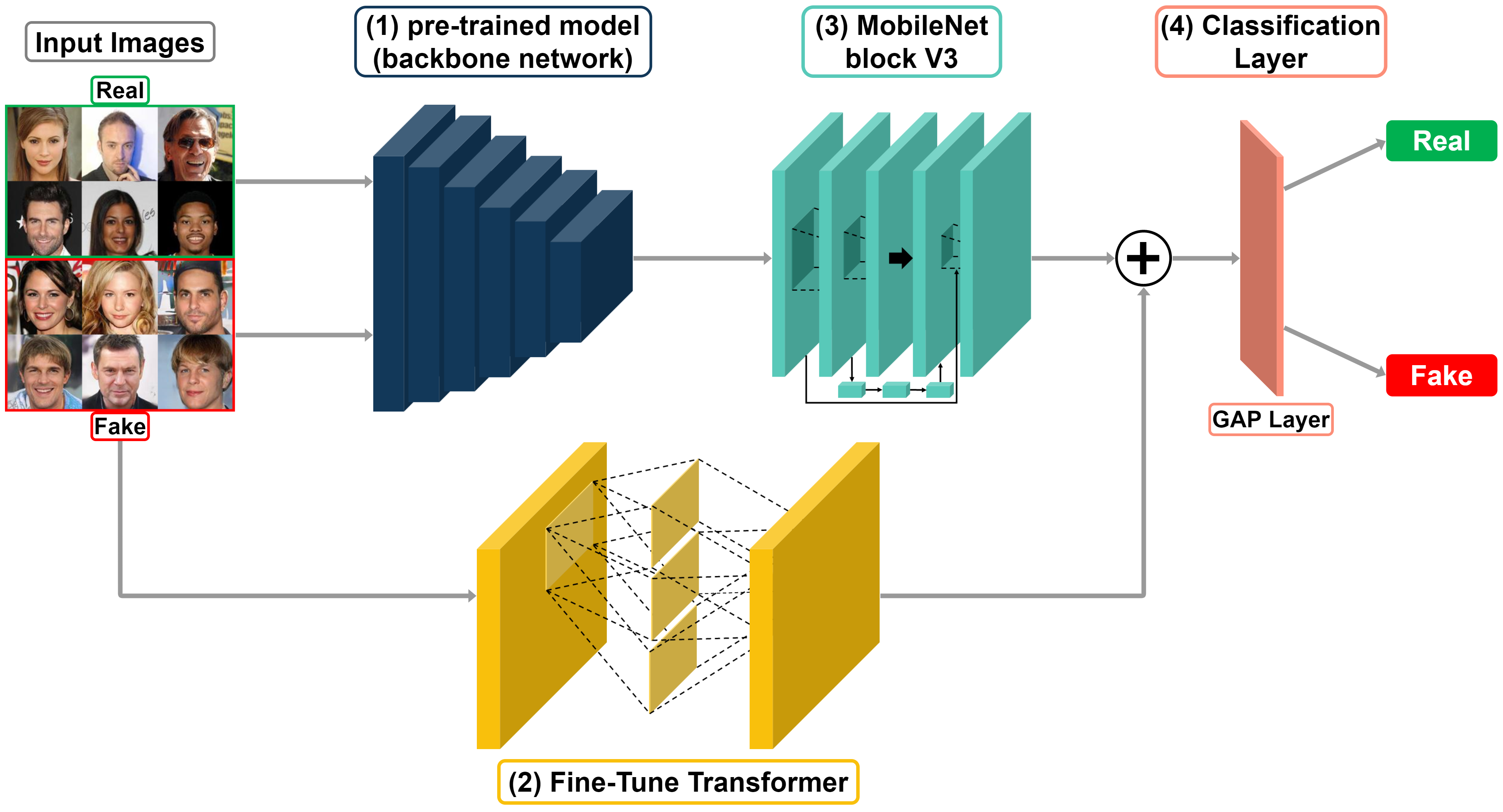}
    \caption{Overview of our \SystemName. \SystemName~modules are shown in yellow and green: (2) Fine-Tune Transformer to an input image and, (3) MobileNet block V3 is attached to (1) pre-trained model (backbone network), where details of each block is shown in Sec. 3. (4) Classification layer, which consists of a global average pooling layer (GAP layer), predicts the real and fake.}
    \label{fig:overview_fdftnet}
\end{figure}

Most of the previous approaches have focused on exploiting metadata information or handcrafted characteristics of images to detect fake images. However, these approaches fail to detect GAN-based fake images, because they are created from scratch and metadata can be also forged; handcrafted features are no longer useful for detection. Recent models, such as \textit{ShallowNet}~\cite{tariq2019gan} and \textit{FakeTalkerDetect}~\cite{jeon2019faketalkerdetect}, used neural networks to detect GANs-generated fake images Yu et al.~\cite{yu2019attributing} used patterns from GAN generated fake to show improvement in detection. \textit{FaceForensics}~\cite{rossler2018faceforensics} showed various forgery detection techniques. However, they lack generalization and will thus have difficulties coping with newly developed DeepFake generation techniques.

In this paper, we propose~\textit{Fake Detection Fine-tuning Network} (\textit{\SystemName}), a new robust fine-tuning neural network-based architecture for fake image detection. ~\SystemName~ combines \textit{Fine-Tune Transformer} (FTT), with a pre-trained Convolutional Neural Network (CNN) as a backbone, and \textit{MobileNet block V3} (MBblockV3). Figure~\ref{fig:overview_fdftnet} shows an overview of our approach, where we utilize well-known, existing CNN architectures~\cite{simonyan2014very,szegedy2015going,he2016deep,iandola2014densenet,hu2018squeeze,howard2017mobilenets} for fake image detection. Our FTT is designed to use different feature extraction from images using the self-attention, and MBblockV3 extracts the feature using different convolution and structure techniques. MBblockV3 is added to the pre-trained model as a backbone network after removing the classification layers. We apply data augmentation by implementing the Cutout method to overcome the limitation of using a small fine-tuning dataset and improve the performance.
Our approach provides a reusable fine-tuning network, improving the existing backbone CNN architectures, which were not designed to detect fake images effectively. Our main contributions are as follows:
\begin{itemize}
    \item We propose \SystemName, a novel neural network-based fake image detector, showing superior performance on detecting fake images compared to previous approaches by achieving 97.02\% accuracy, improving the baseline model accuracy from 4\% to 45\% through our methods.
    \item We provide a robust fine-tuning neural network-based classifier, which requires only a small amount of data for fine-tuning and can be easily integrated with popular CNN architectures.
\end{itemize}

%% file: related.tex
\section{RELATED WORK}
\textbf{Traditional image forgery detection.} Many researchers~\cite{farid2009exposing,krawetz2007picture,yang2015estimating,mankar2015image,yu2019attributing} have investigated various digital forensics algorithms to detect forged images. One way to detect forged images is to analyze them in the frequency domain. However, it is difficult to analyze images with refined, smooth edges, thus giving rise to a different method. In JPEG Ghost~\cite{farid2009exposing}, the forged part is regularly copied from different real images. The normalized pixel distance of the reproduced image differs from the original image, causing a difference in JPEG quality. However, this method will not work if the original image and the forged image have the same quality level. Another approach is Error Level Analysis (ELA)~\cite{krawetz2007picture}, which checks the error level of the images. However, with GANs-generated fake images, ELA cannot classify the error level between the real and generated images.
Another algorithm called the Copy-move Forgery detection~\cite{mankar2015image} is based on Pixel Based approach. Firstly, the dyadic wavelet transform (DWT) is applied to the input image. This transforms the original image to an image of a reduced dimension representation, i.e., the LL1 sub-band. Then this LL1 sub-band is divided into sub-images. To compute the spatial offset between the Copy-move regions, the phase correlation is adopted. The Copy-Move regions can easily be located by pixel matching, which shifts the input image according to the offset and calculates the difference between its shifted version and the original image. In the final step, the Mathematical Morphological Operations (MMO) are used to remove isolated points to improve the location. Traditional digital forensic tools fail to detect GANs-generated images because they are generated as a single image. For this reason, these approaches are not effective.
\newline
\noindent \textbf{Image forgery detection with neural network.} Various CNN-based models have been used to detect forged images. ShallowNet~\cite{tariq2019gan} outperformed previous architectures in detecting real vs. PGGAN with a shallow layer architecture. However, their approach showed limitations when detecting other types of DeepFake images. FaceForensic++~\cite{rossler2019faceforensics++} proposed a forgery detection method tailored to facial manipulations and provided an extensive evaluation in a supervised manner. In addition, they introduced an automatic metric that takes into account the four forms of distortion in realistic scenarios (i.e., random encoding and random dimensions). Using these benchmarks, they analyzed various methods of forgery detection pipeline. However, transfer learning or fine-tuning capabilities were not explored. Recent research by Yu et al.~\cite{yu2019attributing} proposed a method by learning the metadata, mentioned as GAN fingerprints, to effectively detect GANs-generated images. However, our method includes deepfake datasets as well as GANs for detection without the usage of metadata.
\newline
\noindent \textbf{Self-attention and Transformer.} To achieve long-term dependencies on image data, CNN needs to increase the amount of computation via deeper layers, because one-time convolution computation sees only the convolution kernel size. In contrast, self-attention solves this long-term dependency issue by using the softmax outputs of the entire sequence that provide attention to CNN. 
Zhang et al.~\cite{zhang2018self} used self-attention modules to generate images with GANs. Our FTT is different in that we build only self-attention modules, such as Transformer, during the feature extraction in the classification tasks. We apply FTT for the image feature extractor and not for the generator. This approach is similar to the \textit{Multi-head Attention Module}~\cite{vaswani2017attention} (Query, key, and Value), but the difference is that FTT is suitable for the image to be applied to the 1$\times$1 convolution. \newline

%% file: design.tex
\section{Fake Detection Fine-tuning Network (\SystemName)}
% In this section, we describe the architecture design of our \SystemName. The main difference from other fake detection methods is that we utilize well-known, reusable pre-trained models and fine-tune the backbone networks with only a few data to improve the fake detection performance. 
% Figure~\ref{fig:overview_fdftnet} shows an overview of our model, which is composed of 1) a pre-trained model, 2) Fine-Tune Transformer (FTT), and 3) MobileNet V3 blocks (MBblockV3). 

\subsection{Dataset Description}

\begin{figure}[t]
\begin{center}
   \includegraphics[width=0.55\linewidth]{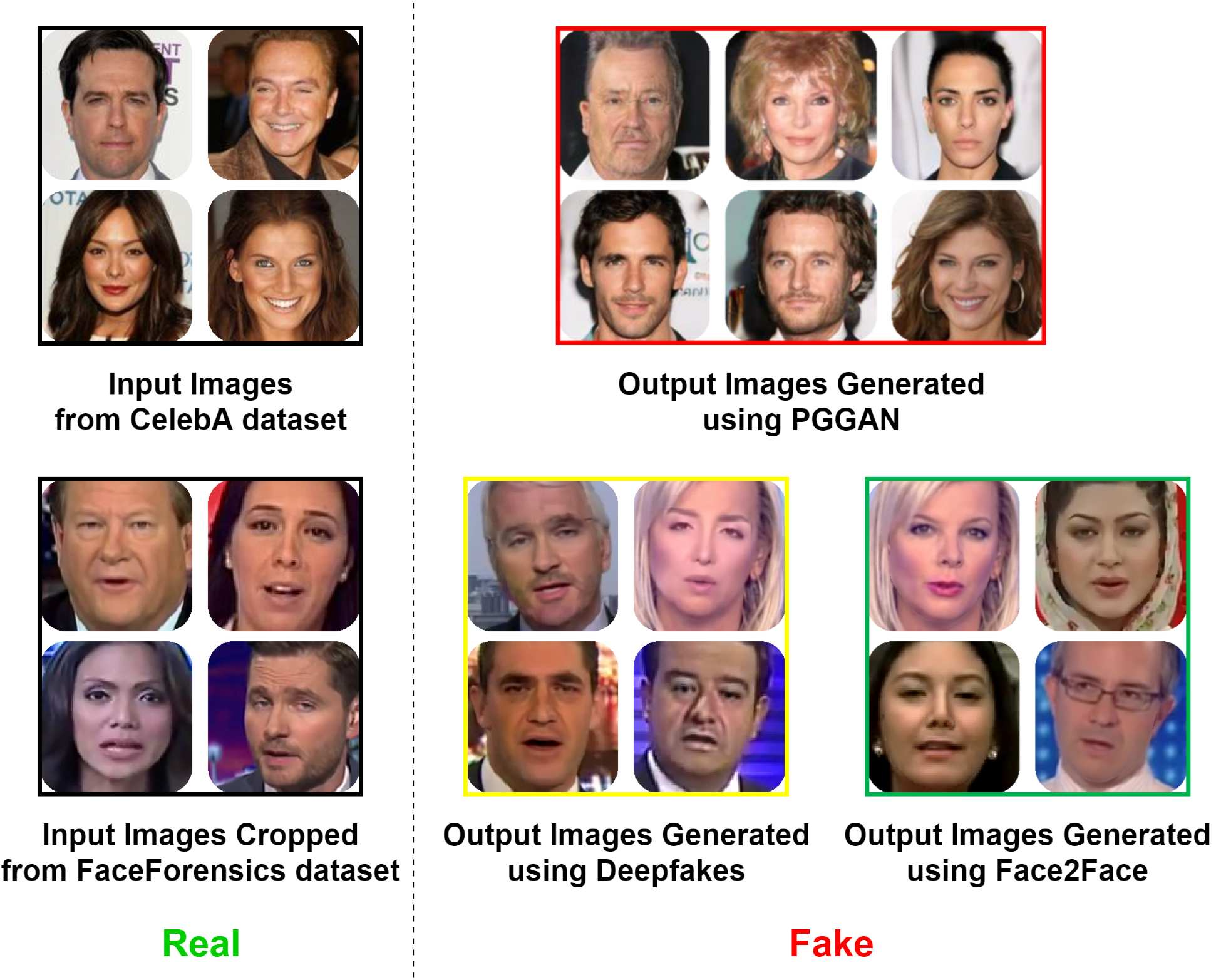}
\end{center}
   \caption{Illustration of our datasets. CelebA~\cite{liu2015deep} images are used as inputs for PGGAN~\cite{karras2017progressive} fake image generation. Images from the FaceForensics~\cite{rossler2018faceforensics} dataset are cropped and used as input images for Deepfakes~\cite{deepfake} and Face2Face~\cite{thies2016face} fake image generation.}
\label{fig:overview}
\end{figure}

\noindent\textbf{CelebA.} CelebFaces Attributes Dataset (CelebA)~\cite{liu2015deep} is a large-scale face attributes dataset with more than 200,000 celebrity images. It is widely used for benchmarking and as inputs for generating training and test datasets for various GAN and VAE approaches. We use CelebA as an input to generate PGGAN~\cite{karras2017progressive} fake images.

\noindent\textbf{PGGAN.} For the GAN-generated image, we used Progressive Growing GANs Dataset (PGGAN)~\cite{karras2017progressive}, consisting of 100,000 GAN-generated fake celebrity images at 1024$\times$1024 resolution using the CelebA dataset. The key idea in PGGAN is to grow both the generator and discriminator progressively. The training starts with both the generator and the discriminator having a low resolution. New layers are added as the training process advances, thus increasing the resolution of the generated images.

\noindent\textbf{Deepfakes.} Deepfakes~\cite{deepfake} was the first publicly available method, which anyone can download and use to produce fake images and videos. The code is based on two autoencoders with a shared encoder. The trained encoder and decoder of the source image are applied to the target image face to produce a forged image. The output of the autoencoder is then blended with the target image. For our experiment, we used the dataset provided by Google/Jigsaw.

\noindent\textbf{FaceForensics.} FaceForensics~\cite{rossler2018faceforensics} is a video dataset comprised of more than 500,000 frames, containing faces from 1004 videos that can be used to study image or video forgeries. An automated version of Face2Face~\cite{thies2016face} approach is used to create the videos. The goal is to animate the facial expressions of the target video by a source actor and re-render the manipulated output video in a photo-realistic fashion. Face2Face re-renders the synthesized target face on top of the corresponding video stream such that it seamlessly blends with the real-world illumination. Since our goal is to detect fake images, we use each frame from the generated output.

\subsection{Description of Pre-trained Backbone CNN networks}
We used the following CNN networks as our backbone networks, as shown in Fig.~\ref{fig:overview_fdftnet}, as well as our baselines (backbone networks): SqueezeNet, ShallowNetV3, ResNetV2, and Xception. Each network is pre-trained from each dataset (i.e., PGGAN, Deepfakes, and Face2Face).

\noindent \textbf{SqueezeNet.} SqueezeNet~\cite{iandola2016squeezenet} has an AlexNet-level accuracy with fewer parameters and would generally have poor performance in fake detection tasks because SqueezeNet is not designed for fake detection. We chose SqueezeNet as the baseline because our~\SystemName~can provide a huge improvement.

\noindent \textbf{ShallowNetV3.} ShallowNetV3~\cite{tariq2019gan} has the highest area under the receiver operating characteristic (AUROC) (93.99\%) on 64$\times$64 resolution images from the CelebA and PGGAN datasets. However, ShallowNetV3 has burdensome fully-connected layers (FC layer) for binary classification. Convolution layers have 115,490 parameters, while FC layers have 4,725,762 parameters. In addition, since this approach has not been tested on deepfakes other than those generated by PGGAN, we aim to investigate the performance.

\noindent \textbf{ResNetV2.}  ResNetV2 has been widely adopted in many image classification tasks. We chose ResNetV2~\cite{he2016identity} as one of the baselines, because ResNetV2 has an opposing characteristic to ShallowNetV3 in terms of the model depth, i.e., ResNetV2 has 50 layers, while ShallowNetV3 has only 8 layers. We believe that these two architectures would show complementary results, and we plan to see the effect of our approach on such deep and shallow CNN architectures.

\noindent \textbf{Xception.} Xception~\cite{chollet2017xception} has been served as the baseline for fake image detection in~\cite{tariq2019gan,rossler2019faceforensics++}. For FaceForenscis++, Xception showed the highest accuracy, i.e., 96.36\% in Deepfake and 86.86\% in Face2Face, justifying our choice of it as a baseline. Xception has no FC layers, but extracts various image feature spaces thanks to \textit{depthwise separable convolutions}, compared to the burdensome FC layers in the ShallowNetV3. We cut the classification layers in a pre-trained model, and add our FTT and MBblockV3 modules.

\subsection{Fine-Tune Transformer (FTT)}
\begin{figure}[t]
\begin{center}
  \includegraphics[width=0.5\linewidth, scale=0.5]{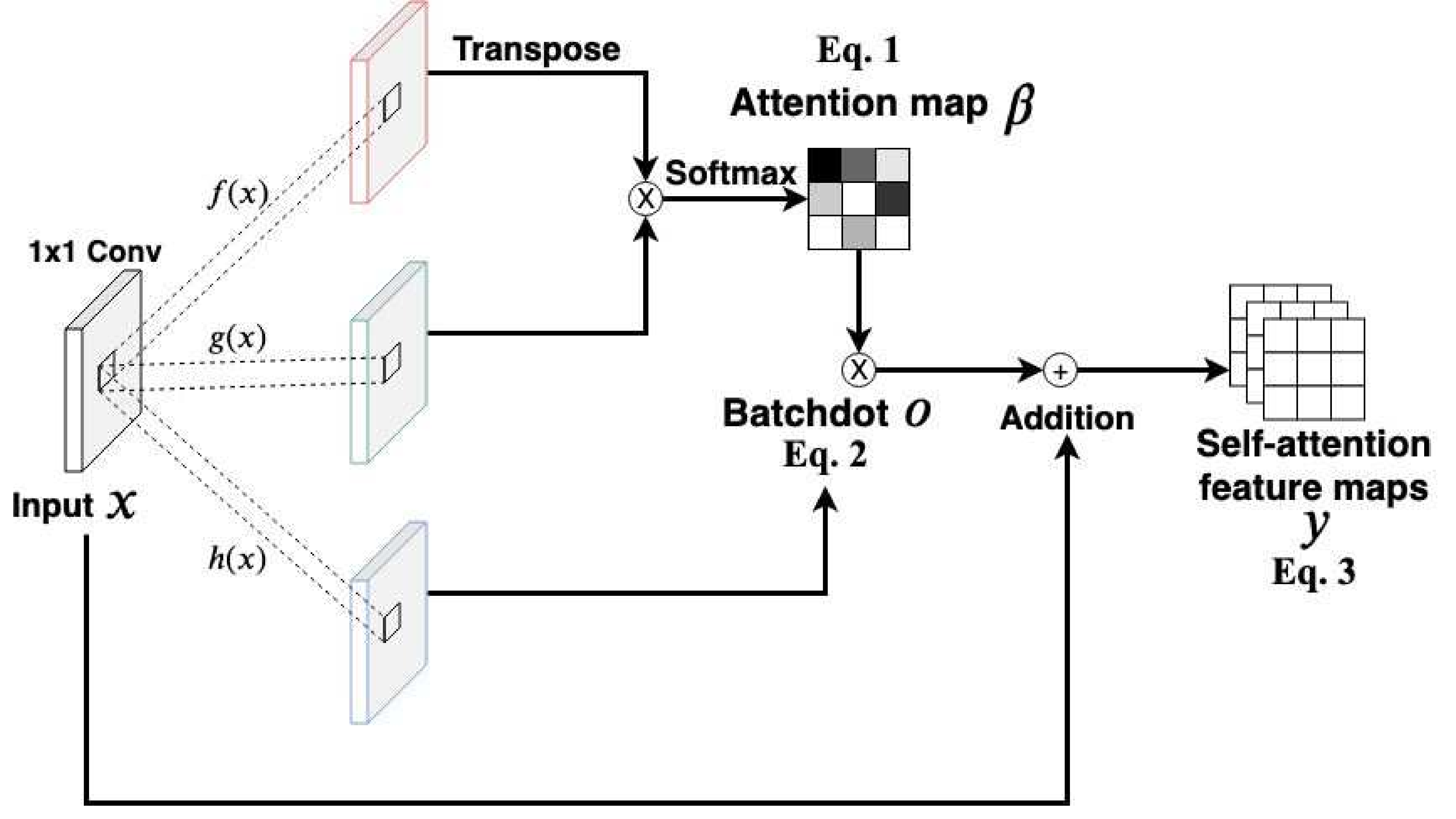}
\end{center}
  \caption{Self-attention module in the Fine-Tune Transformer. The input $x$ (the image or the output from the previous layer) is divided by a $1\times1$ convolution into $f$, $g$, and $h$. The attention map $\beta$ is the softmax output from $f$ and $g$. The batchdot $o$ multiplies $h$ and the attention map $\beta$. The input image $x$ is added to $o$. The final output $y$ is the self-attention feature maps.}
\label{fig:selfattention}
\end{figure}

Fine-Tune Transformer (FTT) consists of several self-attention modules, as shown in Fig.~\ref{fig:selfattention}, where each attention module has $f$($x$), $g$($x$), and $h$($x$) using a 1$\times$1 convolution filter. We iterate $M$ times from the image inputs. $M$ is a hyper-parameter, and we empirically determined that $M$ = 3 yields the highest performance.
\begin{table}[t]
\renewcommand{\arraystretch}{1.0}
\begin{center}
    \scalebox{0.6}{
    \begin{tabular}{ c | c | c | c | c}
    \hline
    Input size & Operation & Num. Parameters & Output dim. & Stride \\
    \hline
    $W\times H\times C$ & 1$\times$1 Conv & 16 & $C$ / $b$ & 1 \\
    $W\times H\times C$ & 1$\times$1 Conv & 16 & $C$ / $b$ & 1 \\
    $W\times H\times C$ & 1$\times$1 Conv & 16 & $C$ & 1 \\
    $W\times H\times C$ / $b$ & Matmul & 0 & $W\times H$ &  - \\
    $WH\times WH$ & Softmax & 0 & $W\times H$ &  - \\
    $WH\times WH$ & Batchdot & 0 & $C$ & - \\
    $W\times H\times C$ & Multiply & 1 & $C$ & -\\
    $W\times H\times C$ & Add & 0 & $C$ & -\\
    \hline
    \end{tabular}}
\end{center}
\caption{Specification for the self-attention module. Conv denotes convolution, $W$, $H$, and $C$ define the input size for the previous layer, and $b$ denotes the bottleneck ratio in the block. The number of parameters are simulated with the following hyperparameters: $W$ = $H$ = $C$ = $64$ and $b$ = $8$.}
\label{table:spec_selfattention}
\end{table}

\begin{table}[t]
\renewcommand{\arraystretch}{0.9}
\begin{center}
    \scalebox{0.55}{
    \begin{tabular}{ c | c | c | c | c}
    \hline
    Input size & Operation & Num. Parameters & Output dim. & Stride \\
    \hline
    $64\times 64\times 3$ & 3$\times$3 DConv & 123 & 32 & 2 \\
    $32\times 32\times 32$ & BN & 128 & 32 & - \\
    $32\times 32\times 32$ & ReLU & 0 & 32 & - \\
    $32\times 32\times 32$ & \textbf{1st Stage self-attention} & 1,321 & \textbf{32} & - \\
    $32\times 32\times 32$ & 3$\times$3 DConv & 2,336 & 64 & 2 \\
    $16\times 16\times 64$ & BN & 256 & 64 & - \\
    $16\times 16\times 64$ & ReLU & 0 & 64 & - \\
    $16\times 16\times 64$ & \textbf{2nd Stage self-attention} & 5,201 & \textbf{64 }& - \\
    $16\times 16\times 64$ & 3$\times$3 DConv & 8,768 & 128 & 2 \\
    $8\times 8\times 128$ & BN & 512 & 128 & - \\
    $8\times 8\times 128$ & ReLU & 0 & 128 & - \\
    $8\times 8\times 128$ & \textbf{3rd Stage self-attention} & 20,641 & \textbf{128} & - \\
    $8\times 8\times 128$ & 1x1 Conv & 73,728 & 576 & 1 \\
    $8\times 8\times 576$ & BN & 2,304 & 576 & - \\
    $8\times 8\times 576$ & ReLU & 0 & 576 & - \\
    $8\times 8\times 576$ & GAP & 0 & 576 & - \\
    \hline
    \end{tabular}}
\end{center}
\caption{Specification for Fine-Tune Transformer (FTT). Conv, BN, DConv, and GAP denote convolution, batch normalization, depth-wise separable convolution, and global average pooling operation, respectively. The ``Attention'' operation in bold indicates the end of one transformer block. We repeat FTT three times ($M$ = $3$) to maximize the performance.
}
\label{table:spec_ftt}
\end{table}

\begin{table}[t]
\renewcommand{\arraystretch}{0.9}
\begin{center}
    \scalebox{0.55}{
    \begin{tabular}{ c | c | c | c | c}
    \hline
    Input size & Operation & Num. Parameters & Output dim. & Stride \\
    \hline
    $W\times H\times C$ & 1$\times$1 Conv & 294,912 & 576 & 1 \\
    $W\times H\times 576$ & BN & 2,304 & 576 & - \\
    $W\times H\times 576$ & h-swish & 0 & 576 & - \\
    $W\times H\times 576$ & 3$\times$3 DConv & 5,184 & 576 & 1 or 2 \\
    $W\times H\times 576$ & BN & 2,304 & 576 &  - \\
    $W\times H\times 576$ & \textbf{GAP} & 0 & 576 & - \\
    $1\times 1\times 576$ & \textbf{1$\times$1 Conv} & 82,944 & 144 & 1 \\
    $1\times 1\times 144$ & \textbf{ReLU} & 0 & 144 & - \\
    $1\times 1\times 144$ & \textbf{1$\times$1 Conv} & 82,944 & 576 & 1 \\
    $1\times 1\times 576$ & \textbf{hard-sigmoid} & 0 & 576 & - \\
    $1\times 1\times 576$ & \textbf{Multiply} & 0 & 576 & - \\
    $W\times H\times 576$ & h-swish & 0 & 576 & - \\
    $W\times H\times 576$ & 1$\times$1 Conv & 73,728 & 128 & 1 \\
    $W\times H\times 128$ & Linear & 0 & 128 & - \\
    $W\times H\times 128$ & BN & 2,304 & 128 & - \\
    $W\times H\times 128$ & Add & 0 & 128 & - \\
    \hline
    \end{tabular}}
\end{center}
\caption{Specification for MBblockV3 with $W$ = $H$ = $8$ and $C$ = $256$. Conv, BN, DConv and GAP denote convolution, batch normalization, depth-wise separable convolution, and global average pooling. $W$, $H$ and $C$ indicate input size. If the stride of 3x3 DConv is 2, the addition operation is skipped, and $W$ and $H$ are divided by 2. Bold operations represent the Squeeze-and-Excitation block.}
\label{table:spec_mbblockv3}
\end{table}

\begin{align}
\begin{split}
    f\left(x\right) &\text{ = } W_f  x, \indent g\left(x\right) \text{ = } W_g  x, \indent h\left(x\right) \text{ = } W_h x ,  \\
    \beta_{j,i} &\text{ = } Softmax\left(f\left(x_i\right)^T  g\left(x_j\right)\right).
    \label{equation:transformer}
\end{split}
\end{align}

In Fig.~\ref{fig:selfattention}, the input $x$ of the previous layers or the input image is divided into three feature spaces $f$($x$), $g$($x$), and $h$($x$). As shown in Eq.~\ref{equation:transformer}, all of them are obtained through the 1$\times$1 convolution, where $W_f, W_g$, and $W_h$ are the respective filter weights of each space. $f$($x$) and $g$($x$) have $b$ channel bottleneck ratio parameter, $\frac{C}{b}$, where $C$ is the number of channels. In this study, we choose $b$ = $8$ as suggested by Zhang et al.~\cite{zhang2018self}. In particular, we use the dot-product attention to produce the attention map $\beta$ in Fig.~\ref{fig:selfattention}, synthesizing the $i^{th}$ and $j^{th}$ locations after the $Softmax$ operation as shown in the above equation.

\begin{align}
    o_j & \text{ = } Batchdot\left(\beta_{j,i} \text{ , } h\left(x\right)\right) \label{equation:batch_dot} \\
    y_i & \text{ = } \gamma o_j \text{ + } x_i. 
    \label{equation:gamma}
\end{align}

\noindent After obtaining the attention map $\beta$, we apply the $Batchdot$ operation to multiply the attention map $\beta_{j,i}$ with $h$($x$), as shown in Eq.~\ref{equation:batch_dot}, and produce output $o_j$. After the $Batchdot$ multiplication, $o_j$ is added to the input $x_i$. Finally, the self-attention feature map, $y_i$, is obtained via multiplying $\gamma$ and adding the input $x_i$, as shown in Eq.~\ref{equation:gamma}. In particular, $\gamma$ is a learnable parameter initialized as 0 at the early stage of learning. This is favorable since the softmax function equally provides attention to all the feature spaces at the early stage of learning.

Next, in our FTT, we apply the self-attention module three times ($M$ = 3) with an input size of 64$\times$64$\times$3, as shown in Table~\ref{table:spec_ftt}. The first layer is a 3$\times$3 separable convolution with 32 filters and 2 strides followed by Batch Normalization (BN)~\cite{ioffe2015batch} and ReLU. The dimension of the output feature map from the self-attention module is 32, 64, and 128, respectively; the width (the number of channels) is doubled when the resolution is down-sampled, as shown in Table~\ref{table:spec_ftt}. After that, self-attention is performed three times ($M$ = $3$), followed by SeparableConv3$\times$3, BN, and ReLU. The main reason we apply self-attention modules in FTT is to overcome the limitations of CNN in achieving long-term dependencies, caused by the use of numerous Conv filters with a small size.
On the other hand, only one-time use of the FTT is necessary to achieve the long-term dependencies, avoiding the construction of deep CNN layers. Also, a three-time application of self-attention modules allows us to explore and learn diverse deep features of the input images via fine-tuning.

\subsection{MobileNet block V3}
We chose MobileNet block V3 (MBblockV3) to explore the image feature space through inverted residual structure and linear bottleneck~\cite{sandler2018mobilenetv2}. Depthwise separable convolutions, as in Xception and MobileNetV1 \cite{howard2017mobilenets}, are also included in MBblockV3. Comprehensively, MobileNet is an architecture that has already proven its efficiency by using a small number of parameters, drastically increasing computational efficiency. We chose MBblockV3, because it is a suitable module for the efficient extraction of the feature space over the pre-trained feature space. FTT and MBblockV3 are repeatedly used $M$ and $N$ times, respectively. Each of them is added before the final classification layer. MBblockV3 has the parameter $N$ after the pre-trained model. In our experiment, we use $N$ = 4, determined empirically, yielding the best performance for fine-tuning.
In particular, we use the modified \textit{h-swish}~\cite{howard2019searching} and the ReLU6 as activation functions. This non-linearity~\cite{ramachandran2017swish,elfwing2018sigmoid,hendrycks2016bridging} significantly improves the performance of neural networks and is defined as follows:
\begin{equation}
    \label{equation1}
        \text{h-swish[x]} \text{ = } x\frac{\text{ReLU6} \left (x \text{ + } 3 \right)}{6}, 
        \text{ where }
        \text{ReLU6[x]} \text{ = } min\left(max\left(0, x\right), 6 \right).
\end{equation}
\indent Since clipping the input values at the bottom layers may have a side effect of distorting the data distribution~\cite{sheng2018quantization}, we apply these activation functions at the top layers to reduce distortion and extract different signals from ReLU.
Next, the \textit{Squeeze-and-Excitation blocks} (SE block) in Squeeze and Excitation networks~\cite{hu2018squeeze} are applied in the bottleneck layer. Global information on the image resolution is embedded in the squeeze stage, and information aggregation is used to capture channel dependencies and is re-calibrated through the gated computation (element-wise multiplication), similar to the attention mechanism in the excitation stage. Details of the SE block parameters are summarized in Table~\ref{table:spec_mbblockv3}.

%% file: result.tex
\section{Experimental Results}

\subsection{Training details}
\begin{table}[t]
\renewcommand{\arraystretch}{0.4}
\begin{center}
    \scalebox{0.6}{
    \begin{tabular}{l |c|c|c|c}
        \hline
         \\ Dataset & Train & Validation & Test & \textbf{Fine-tune} \\
        \hline
         \\ PGGAN & 128,404 & 32,100 & 37,566 & \textbf{1,000 (real), 1,000 (fake)} \\
        \hline
         \\ Deepfake & 60,000 & 18,000 & 20,000 & \textbf{1,000 (real), 1,000 (fake)} \\
        \hline
         \\ Face2Face & 60,000 & 18,000 & 20,000 & \textbf{1,000 (real), 1,000 (fake)} \\
         \hline
    \end{tabular}}
\end{center}
\caption{The respective size of the train, validation, test, and fine-tune sets. We use only 1,000 real and fake images, respectively, for fine-tuning.}
\label{table:volume}
\end{table}
\begin{figure}[t]
\centering
    \includegraphics[width=0.45\linewidth]{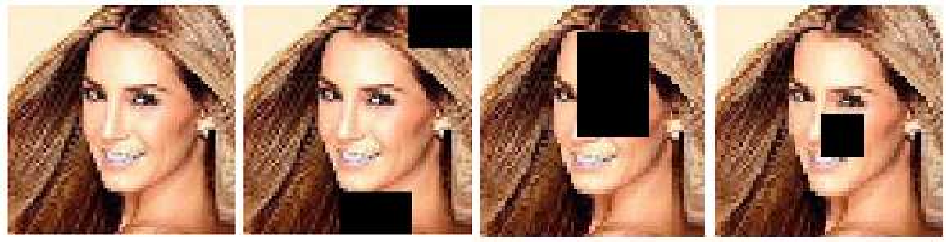}
\caption{Example of a Cutout data augmentation. Random regions of the original image (left) are masked out by black rectangles. Every epoch, the rectangular mask changes in form and all images are resized to 64$\times$64 resolution.}
\label{fig:cutout}
\end{figure}
All datasets have train, validation, test, and fine-tune sets. The size of each dataset is shown in Table~\ref{table:volume}. Our \SystemName~is trained with Stochastic Gradient Descent (SGD) with momentum for 300 epochs on all datasets. The learning rate is initialized at 0.3 and annealed using a cosine function. The momentum rate is set to 0.9, and the mini-batch size is set to 128. Early stopping is applied, when the validation loss ceases to decrease for 20 epochs. To reenact the most challenging scenario in detecting fake images, all input images are resized to 64$\times$64 resolution.

\noindent\textbf{Data augmentation.} Input images are translated into a width and height range of [-2, 2] with the nearest-padding on empty pixels generated after translation. Zoom and rotation are also applied to a degree range of [-0.2, 0.2]. We also perform random horizontal flipping. These data augmentations are applied to all fine-tune sets. For validation and test sets, only a $1$ / $255$ scaling augmentation to the input image is applied.

\noindent\textbf{Cutout.} Cutout method applies squared zero masks on a random location of each input image. Fig.~\ref{fig:cutout} presents an example of a Cutout data augmentation. DeVries et al.~\cite{devries2017improved} used random zero masks of 16 pixels for CIFAR-10 (32$\times$32 pixels images), 5 random iteration parameters $\alpha$ for cutting, and 16 random size multipliers $\beta$ for the cutting masks. We use 4$\times$4 pixels mask, 3 iterations, and 5-size multipliers for cutting masks for 64$\times$64 images ($\alpha$ = 3 and $\beta$ = 5). Since we use random translation, we do not use random center cropping, which was used in the original paper. When we conducted with the original setting, we faced severe underfitting with no convergence of losses. We observed higher performance with a setting of low Cutout parameters  ($\alpha$ = 3 and $\beta$ = 5) as compared to the implementation without Cutout, which showed strong overfitting. Because we fine-tune with a small amount of data, we apply this non-aggressive parameter setting.
\subsection{Performance evaluation}
\begin{table}[t]
\begin{center}
\renewcommand{\arraystretch}{1.0}
    \scalebox{0.7}{
    \begin{tabular}{l||c||cc||cc||cc}
    \hline
    \multirow{2}{*}{Model} & \multirow{2}{*}\ Dataset & \multicolumn{2}{c}{PGGAN} & \multicolumn{2}{c}{Deepfake} & \multicolumn{2}{c}{Face2Face} \\\cline{2-8} & Backbone & ACC (\%) & AUROC & ACC (\%) & AUROC  & ACC (\%) & AUROC  \\
    \hline
    SqueezeNet & baseline & 50.00 & 50.00 & 50.00 & 50.00 & 50.00 & 50.00 \\
    \SystemName \space (Ours) & SqueezeNet & \underline{88.89} & \underline{92.76} & \underline{92.82} & \underline{97.61} & \underline{87.73} & \underline{94.20} \\
    \hline
    ShallowNetV3$\dagger$& baseline & 85.73 & 92.90 & 89.77 & 92.81 & 83.35 & 88.49 \\
    \SystemName \space (Ours) & ShallowNetV3 & \underline{88.03} & \underline{94.53} & \underline{94.29} & \underline{97.83} & \underline{84.55} & \underline{93.28} \\
    \hline
    ResNetV2 & baseline & 84.80 & 88.58 & 81.52 & 89.72 & 58.83 & 62.47 \\
    \SystemName \space (Ours) & ResNetV2 & \underline{84.83} & \underline{94.05} & \underline{91.03} & \underline{96.08} & \underline{85.15} & \underline{92.91} \\
    \hline
    Xception & baseline & 87.12 & 94.96 & 95.10 & 98.92 & 85.78 & 93.67 \\
    \SystemName \space (Ours) & Xception & \textbf{90.29} & \textbf{95.98}  & \textbf{97.02} & \textbf{99.37} &  \textbf{96.67} & \textbf{98.23} \\
    \hline
    \end{tabular}}
    \end{center}
\caption{Overall performance evaluation results. The evaluation metrics used are ACC (\%) and AUROC (\%). The underlined results are improved performance compared to the baseline and the best detection results among all are highlighted in bold.} 
\label{table:performance}
\end{table}
We present our overall performance results in Table~\ref{table:performance}. In Table~\ref{table:performance}, we use the accuracy (ACC) and AUROC as evaluation metrics. We experimented with all four baseline models on each dataset with similar training strategies. The experimental results show that our \SystemName~has superior detection performance in both ACC and AUROC, compared to all the baselines. In terms of training data size, our model shows high performance using 1,000 images for real and fake, respectively.

\noindent\textbf{PGGAN.} To yield the best detection performance, we freeze the weight parameters of all layers of the pre-trained models. FTT with parameter $M$ = 3 is used, and MBblockV3 with parameter $N$ = 2 is added; the same data augmentation is applied. Table~\ref{table:performance} shows the results of our models compared with the baseline models. Our results show that Xception, among all baseline models, achieved the highest performance (87.12\% ACC and 94.96\% AUROC). Our model showed a performance of 90.29\% ACC and 95.98\% AUROC, which is higher than that of ShallowNetV3 with an ensemble~\cite{tariq2019gan}. ShallowNetV3 is improved from 85.73\% and 92.90\% ACC to 88.03\% and 94.53\% AUROC, respectively, similar to the ensemble version. SqueezeNet baseline shows the lowest baseline performance, but it is significantly improved to a similar level to that of ShallowNetV3, from 50.00\% to 92.76\%, by applying our model.
\vspace{1mm}

\noindent\textbf{Deepfake.} 
Here also, the same data augmentation techniques are applied. For FTT, we use $M$ = $3$ and $N$ = $4$ for MBblockV3. Cutout has $\alpha$ = $3$ iteration parameters and $\beta$ = 10 multiplier parameters. The results show that all models achieve significant improvement in performance. Table~\ref{table:performance} indicates that Xception has the highest performance of 95.10\% ACC and 98.92\% AUROC. Using our approach, this baseline model is also improved to 97.02\% ACC and 99.37\% AUROC. ShallowNetV3 has 89.77\% ACC and 92.81\% AUROC. They increased to 94.29\% ACC and 97.83\% AUROC, respectively. ResNetV2 is also improved from 81.52\% ACC and 89.72\% AUROC to 91.03\% ACC and 96.08\% AUROC. SqueezeNet baseline shows the lowest performance, 50.00\% ACC and AUROC, but is improved to 92.82\% ACC and 97.61\% AUROC.  
\vspace{1mm}

\noindent\textbf{Face2Face.} The training strategies for Face2Face are very similar to those of the Deepfake dataset. Data augmentation is also applied. $M$, $N$, $\alpha$, and $\beta$ are set to $3$, $4$, $3$, and $10$, respectively.
The interesting point is that ResNetV2 baseline performed poorly (58.83\% ACC and 62.47\% AUROC), but significant improvements are made using our methods (85.15\% ACC and 92.91\% AUROC). Our results demonstrate the generalization ability of our approach, improving the poorly performing baseline above 90\% across all models and datasets. Compared to \textit{FaceForensics Benchmark Results}~\cite{faceforencis}, the highest state-of-the-art method is Xception, which shows 96.4\% ACC in Deepfake and 86.9\% ACC in Face2Face. Our~\SystemName~achieves higher performance (97.02\% and 96.67\%) than the current state-of-the-art method for the same dataset.

%% file: ablation.tex
\section{Ablation study, discussions, and limitations}
\begin{table}[t]
\renewcommand{\arraystretch}{0.4}
\begin{center}
\scalebox{0.7}{
\begin{tabular}{l |c|c|c|c}
    \hline
    \\ Method & backbone & Dataset & Acc & AUROC \\
    \hline
    \\ With FTT & Xception & Deepfake & \textbf{97.02}\% & \textbf{99.37}\% \\
    \\ Without FTT & Xception & Deepfake & 94.56\% & 98.89\% \\
    \hline
\end{tabular}}
\end{center}
\caption{Ablation study for Fine-Tune Transformer (FTT). Our model with FTT has 2.46\% higher accuracy (ACC) than those without FTT, increasing the ACC from 94.56\% to 97.02\%.}
\label{table:perform_ftt}
\end{table}
In this section, we validate each module and technique through an ablation study. In Table~\ref{table:perform_ftt}, we choose the Xception model and the Deepfake dataset to compare our model with and without the FTT, while all other settings remain the same. With FTT, we can achieve about 2.5\% higher performance than without FTT, as shown in Table~\ref{table:perform_ftt}.
Our current work has the following limitations: First, we used both real and fake data for training and fine-tuning, but we have constrained resources in practice. In FakeTalkerDetect~\cite{jeon2019faketalkerdetect} for fake detection, researchers used Siamese networks for training only on real data. However, in our implementation, few-shot learning and unbalanced learning are major obstacles to achieving high performance. Second, transfer learning is required to improve performance. We trained each model on each dataset. For future work, we plan to research the transfer learning ability to further generalize our model.

%% file: conc.tex
\section{Conclusion}
We propose~\SystemName, which is a robust fine-tuning neural network-based architecture, to detect fake images and significantly improve the baseline CNN architectures. Our model achieves the state-of-the-art accuracy in fake image detection on the GAN-based dataset and the Deepfake-based dataset. Our experimental results with the use of a limited amount of data show the exploration and exploitation of image feature space beyond the pre-trained models. Our results show that \SystemName~is a promising method for detecting fake images generated by powerful deep learning methods, requiring only a small amount of images for re-training. Therefore, \SystemName~can be a viable option even for detecting new fake images in a real-world scenario, where available datasets are extremely limited. Further, we offer open source versions of our work for it to be widely leveraged by the research community\footnote{\url{https://github.com/cutz-j/FDFtNet}}.

%% file: ack.tex
\section*{Acknowledgements}
We thank Siho Han for providing his expertise to greatly improve this work. This work was partly supported by Institute of Information  communications Technology Planning \& Evaluation (IITP) grant funded by the Korea government (MSIT) (No.2019-0-00421, AI Graduate School Support Program (Sungkyunkwan University)). Also, this research was supported by Energy Cloud R\&D Program through the National Research Foundation (NRF) of Korea funded by the Ministry of Science, ICT (No. 2019M3F2A1072217), and was supported by the National Research Foundation of Korea (NRF) grant funded by the Korea government (MSIT) (No. 2017R1C1B5076474, and 2020R1C1C1006004).